\documentclass[runningheads]{llncs}
\usepackage{graphicx}
\usepackage{amsmath}
\usepackage{amsmath,amssymb,amsfonts}
\usepackage{graphicx}
\usepackage{textcomp}
\usepackage{xcolor}

\usepackage{algorithm}
\usepackage{algorithmic}
\usepackage{booktabs}

\usepackage{caption}

\usepackage[T1]{fontenc} 

\usepackage{epstopdf}

\DeclareMathOperator{\mape}{\textsc{mape}}
\DeclareMathOperator{\mpe}{\textsc{mpe}}
\DeclareMathOperator{\ape}{\textsc{ape}}
\DeclareMathOperator{\pe}{\textsc{pe}}
\DeclareMathOperator{\rmse}{\textsc{rmse}}

\begin{document}
\title{Randomized Neural Networks for Forecasting Time Series with Multiple Seasonality\thanks{Supported by Grant 2017/27/B/ST6/01804 from the National Science Centre, Poland.}}
\titlerunning{Randomized NNs for Forecasting Time Series with Multiple Seasonality}
%
\author{Grzegorz Dudek\orcidID{0000-0002-2285-0327}}
\authorrunning{G. Dudek}
%
\institute{Electrical Engineering Faculty, Częstochowa University of Technology,\\ Częstochowa, Poland\\
\email{grzegorz.dudek@pcz.pl}}
\maketitle              
\begin{abstract}

This work contributes to the development of neural forecasting models with novel randomization-based learning methods. These methods improve the fitting abilities of the neural model, in comparison to the standard method, by generating network parameters in accordance with the data and target function features. A pattern-based representation of time series makes the proposed approach useful for forecasting time series with multiple seasonality. In the simulation study, we evaluate the performance of the proposed models and find that they can compete in terms of forecasting accuracy with fully-trained networks.  
Extremely fast and easy training, simple architecture, ease of implementation, high accuracy as well as dealing with nonstationarity and multiple seasonality in time series make the proposed model very attractive for a wide range of complex time series forecasting problems.

\end{abstract}

\section{Introduction}

Time series (TS) expressing different phenomena and processes may include multiple seasonal cycles of different lengths. They can be observed in demand variations for various goods, weather conditions, customer numbers, stock market indicators or results of experimental research. Multiple seasonality in TS as well as nonstationarity, nonlinear trend and random fluctuations place high demands on forecasting models. The model should be flexible enough to capture these features without imposing too much computational burden. Over the years, many sophisticated forecasting models for TS with multiple seasonality have been proposed including statistical and machine learning (ML) ones.  

One of the most commonly employed classical approaches, the autoregressive moving average model (ARMA), can be extended to multiple seasonal cycles by including additional seasonal factors \cite{Box94}.
Another popular statistical model,  Holt–Winters exponential smoothing (ETS), was developed for forecasting TS data that exhibits both a trend and a seasonal variation. ETS was extended to incorporate a second and a third seasonal component in \cite{Taylor10}. Both these models, ARMA and seasonal Holt–Winters model, have a significant weakness. They require the same cyclical behavior for each period. In \cite{Gould08}, to cope with changing seasonal patterns, innovations state space models for ETS were proposed. The limitation of the model is that it can only be used for double seasonality where one seasonal length is a multiple of the other. A further extension of ETS was proposed in \cite{DeLivera11}. To deal with multiple seasonal periods, high-frequency, non-integer seasonality, and dual-calendar effects, it combines an ETS state space model with Fourier terms, a Box-Cox transformation and ARMA error correction.  

As an alternative to statistical models, ML models have the ability to learn relationships between predictors and forecasted variables from historical data. One of the most popular in the well-stocked arsenal of ML methods are neural networks (NNs). A huge number of forecasting models based on different NN architectures have been proposed \cite{Benidis20}. They deal with multiple seasonality differently, depending on the specific architectural features and the creativity of the authors. For example, the model that won the renowned M4 Makridakis competition combines ETS and recurrent NN (RNN) \cite{Smyl20}. In this approach, ETS produces two seasonal components for TS deseasonalization and adaptive normalization during on-the-fly preprocessing, while RNN, i.e. long short term memory (LSTM), predicts the preprocessed TS.      

Another example of using LSTM for forecasting TS with multiple seasonal patterns was proposed recently in \cite{Bandara20}. To deal with multiple seasonal cycles, the model initially deseasonalizes TS using different strategies including Fourier transformation. RNNs, such as LSTM, gated recurrent units, and DeepAR \cite{Sal20}, dominate today as NN architectures for TS forecasting thanks to their powerful ability to process sequential data and capture long-term dependencies. But other deep architectures are also useful for forecasting multiple seasonal TS. For example, N-Beats \cite{Oresh20} was designed specifically for TS with multiple seasonality. It is distinguished by a specific architecture including backward and forward residual links and a very deep stack of fully-connected layers. 

The above presented approaches to forecasting TS with multiple seasonal periods rely on incorporating into the model mechanisms which allow it to deal with seasonal components. This complicates the model and makes it difficult to train and optimize. An alternative approach is to simplify the forecasting problem by TS decomposition or preprocessing. In \cite{Dudek16}, TS with three seasonal cycles was represented by patterns expressing unified shapes of the basic cycle. This preprocessing simplified the relationship between TS elements, making decomposition unnecessary and removing the need to build a complex model. Instead, simple shallow NNs can be used \cite{Dudek16} or nonparametric regression models \cite{Dudek15}. Experimental research has confirmed that these models can compete in terms of accuracy with state-of-the-art deep learning models, like the winning M4 submission \cite{Dudek20}.     

In this study, we use a pattern representation of TS to simplify the forecasting problem with multiple seasonality and propose randomization-based shallow NNs to solve it. Randomized learning was proposed as an alternative to gradient-based learning as the latter is known to be time-consuming, sensitive to the initial parameter values and unable to cope with the local minima of the loss function. In randomized learning, the parameters of the hidden nodes are selected randomly and stay fixed. Only the output weights are learned. This makes the optimization problem convex and allows us to solve it without tedious gradient descent backpropagation, but using a standard linear least-squares method instead \cite{Pri15}. This leads to a very fast training. The main problem in randomized learning is how to select the random parameters to ensure the high performance of the NN \cite{Cao18}, \cite{Zha16}. In this study, to generate the random parameters we use three methods recently proposed in \cite{Dud20a}, \cite{Dud20}. These methods distribute the activation functions (sigmoids) of hidden nodes randomly in the input space and adjust their weights (or a weight interval) to the target function (TF) complexity using different approaches.    

The main goal of this study is to show that randomization-based NNs can compete in terms of forecasting accuracy with fully-trained NNs. The contribution of this study can be summarized as follows:
\begin{enumerate}
	\item A new forecasting model for TS with multiple seasonality based on randomized NNs is proposed. To deal with multiple seasonality and nonstationarity, the model applies pattern representation of TS in order to simplify the relationship between input and output data.
	\item Three randomization-based methods are used to generate the NN hidden node parameters. They introduce steep fragments of sigmoids in the input space, which improves modeling of highly nonlinear TFs. A randomized approach leads to extremely fast and easy training, simple NN architecture and ease of implementation.
	\item Numerical experiments on several real-world datasets demonstrate the efficiency of the proposed randomization-based models when compared to fully-trained NNs.
\end{enumerate}

The remainder of this work is structured as follows. Section 2 presents the proposed forecasting model based on randomized NNs, and a TS representation using patterns of seasonal cycles and three methods of generating NN parameters are described. The performance of the proposed approach is evaluated in Section 3. Finally, Section 4 concludes the work.

\section{Forecasting model}

\begin{figure}
\centering
\includegraphics[width=0.95\textwidth]{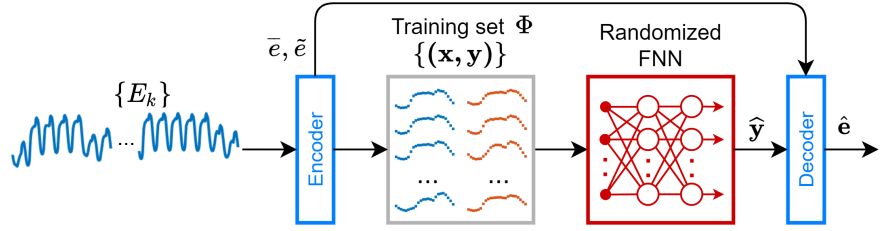}
\caption{Block diagram of the proposed forecasting model.} \label{figModel}
\end{figure}

The proposed forecasting model is shown in Fig. \ref{figModel}. It is composed of encoder and decoder modules and a randomized feedforward NN (FNN). The model architecture, its specific features, and components are described below.

\subsection{Encoder}

The task of the encoder is to convert an original TS into unified input and output patterns of its seasonal cycles. To create input patterns, the TS expressing multiple seasonality, $\{E_k\}_{k=1}^K$, is divided into seasonal sequences of the shortest length. Let these sequences be expressed by vectors $ \mathbf{e}_i = [E_{i,1}, E_{i,2}, …, E_{i,n}]^T $, where $n$ is the seasonal sequence length and $i=1, 2, ..., K/n$ is the sequence number. These sequences are encoded in input patterns $\mathbf{x}_i = [x_{i,1}, x_{i,2}, …, x_{i,n}]^T$ as follows:

\begin{equation}\label{eq1}
\mathbf{x}_i = \frac{\mathbf{e}_i-\overline{{e}}_i}{\widetilde{e}_i}
\end{equation}
where $\overline{{e}}_i$ is a mean value of sequence $\mathbf{e}_i$, and $\widetilde{e}_i = \sqrt{\sum_{t=1}^{n} (E_{i,t}-\overline{{e}}_i)^2}$ is a measure of sequence $\mathbf{e}_i$ dispersion.

\begin{figure}[h]
\centering
\includegraphics[width=0.48\textwidth]{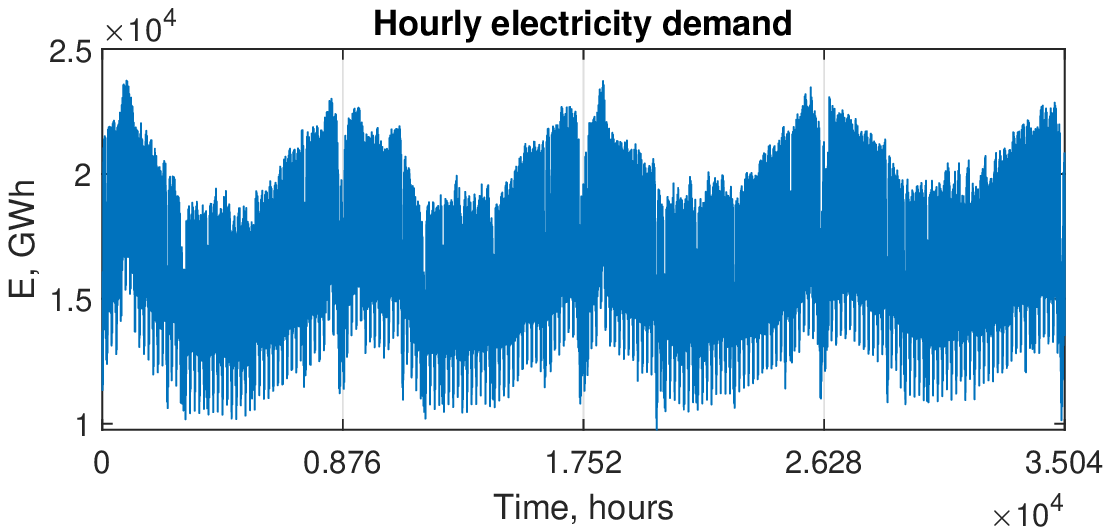}
\includegraphics[width=0.48\textwidth]{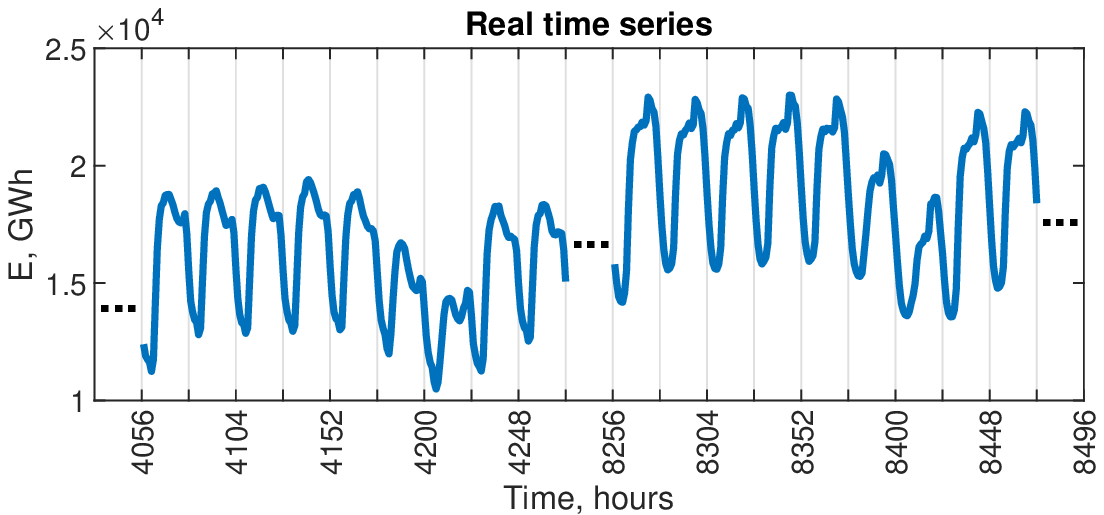}

\includegraphics[width=0.48\textwidth]{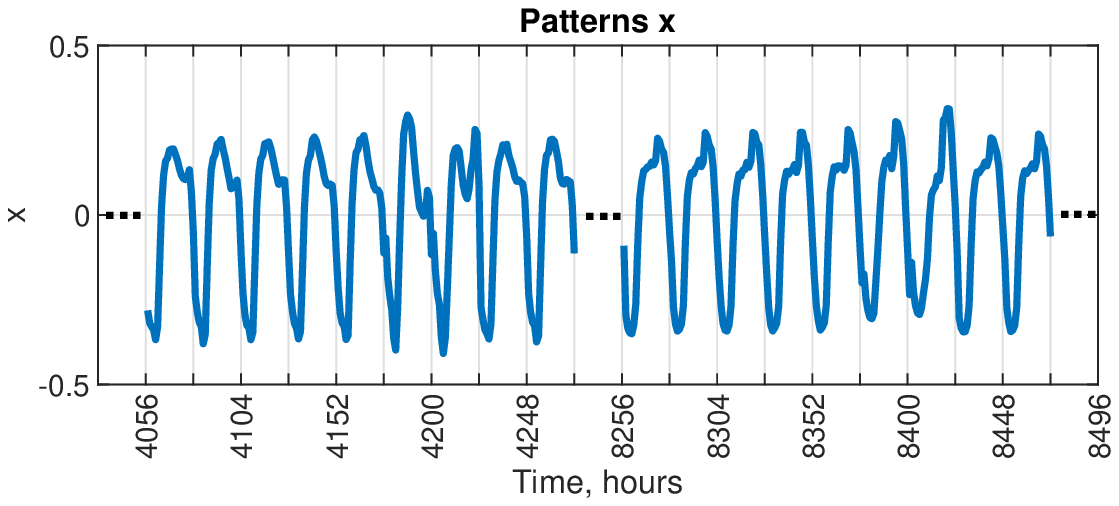}
\includegraphics[width=0.48\textwidth]{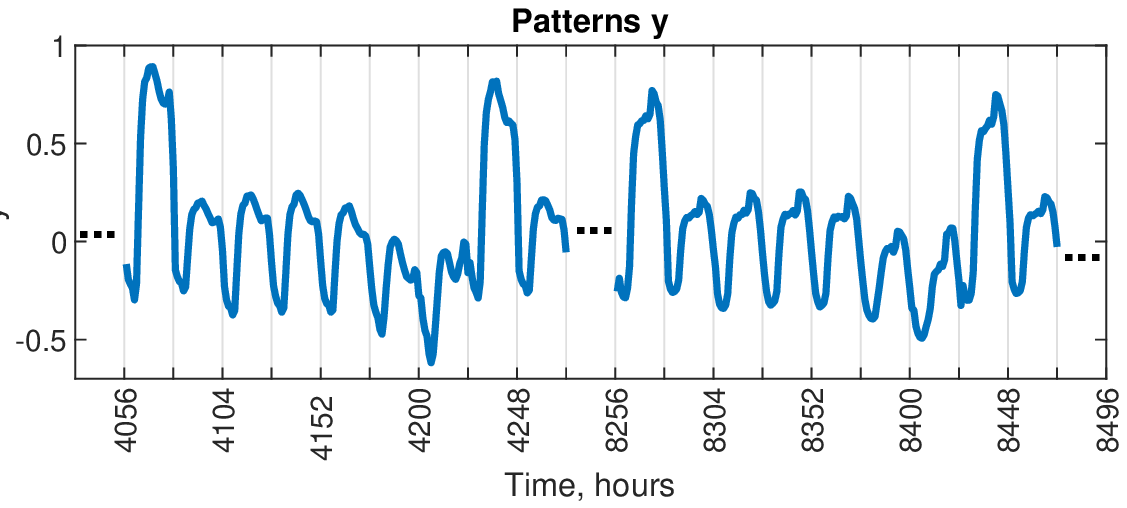}
\caption{Real hourly electricity demand TS and its x- and y-patterns.} \label{figTS}
\end{figure}

Note that the x-patterns are normalized versions of centered vectors $\mathbf{e}_i$. All x-patterns, representing successive seasonal sequences, have zero mean, the same variance and the same unity length. However, they differ in shape. Thus, the original seasonal sequences, which have a different mean value and dispersion, are unified. This is shown in Fig. \ref{figTS} on the example of the hourly electricity demand TS expressing three seasonalities: daily, weekly, and yearly. Note that the x-patterns representing the daily cycles are all normalized and differ only in shape.

The output patterns $\mathbf{y}_i = [y_{i,1}, y_{i,2}, …, y_{i,n}]^T$ represent the forecasted sequences $ \mathbf{e}_{i+\tau} = [E_{i+\tau,1}, E_{i+\tau,2}, …, E_{i+\tau,n}]^T $, where $\tau \geq 1$ is a forecast horizon. The y-patterns are determined as follows:

\begin{equation}\label{eq2}
\mathbf{y}_i = \frac{\mathbf{e}_{i+\tau}-\overline{{e}}_i}{\widetilde{{e}}_i}
\end{equation}
where $\overline{{e}}_i$ and $\widetilde{e}_i$ are the same as in \eqref{eq1}.

Note that in \eqref{eq2}, for the $i$-th output pattern, we use the same coding variables $\overline{{e}}_i$ and $\widetilde{e}_i$ as for the $i$-th input pattern. This is because the coding variables for the forecasted sequence, $\overline{{e}}_{i+\tau}$ and $\widetilde{e}_{i+\tau}$, are unknown for the future period. Using the coding variables determined from the previous period has consequences which are demonstrated in Fig. \ref{figTS}. Note that y-patterns in this figure reveal the weekly seasonality. The y-patterns of Mondays are much higher than the patterns of other days of the week because the Monday sequences are coded with the means of Sunday sequences which are much lower than the means of Monday sequences. For similar reasons, y-patterns for Saturdays and Sundays are lower than y-patterns for the other days of the week. Thus, the y-patterns are not unified globally but are unified in groups composed of the same days of the week. For this reason, we construct forecasting models that learn from data representing the same days of the week. For example, when we train the model to forecast the daily sequence for Monday, the training set for it, $\Phi=\{(\mathbf{x}_i,\mathbf{y}_i)\}_{i=1}^N$, is composed of the y-patterns representing all Mondays from history and corresponding x-patterns representing the previous days (depending on the forecast horizon; Sundays for $\tau = 1$).

\subsection{Decoder}

The decoder converts a forecasted output pattern into a TS seasonal cycle. The output pattern  predicted by randomized FNN is decoded using the coding variables of the input query pattern, $\mathbf{x}$, using transformed equation \eqref{eq2}:

\begin{equation}\label{eq3}
\widehat{\mathbf{e}} = \widehat{\mathbf{y}}\widetilde{{e}}+\overline{{e}} 
\end{equation}
where $\widehat{\mathbf{e}}$ is the forecasted seasonal sequence, $\widehat{\mathbf{y}}$ is the forecasted output pattern, $\widetilde{e}$ and $\overline{e}$ are the coding variables determined from the TS sequence encoded in query pattern $\mathbf{x}$.

\subsection{Randomized FNN}

The randomized FNN is composed of $n$ inputs, one hidden layer with $m$ nonlinear nodes, and $n$ outputs. Logistic sigmoid activation functions are employed for hidden nodes. 
The training set is $ \Phi = \left\lbrace (\mathbf{x}_i, \mathbf{y}_i)\right\rbrace_{i=1}^N, \mathbf{x}_i, \mathbf{y}_i \in \mathbb{R}^n$. The randomized learning algorithm consists of three steps \cite{Dud19}.
\begin{enumerate}
  \item Randomly generate hidden node parameters, i.e. weights $ \mathbf{a}_j = [ a_{j,1}, a_{j,2}, \ldots,\\ a_{j,n}]^T $ and biases $ b_j, j = 1, 2, \ldots, m $, according to any continuous sampling distribution. 

  \item Calculate the hidden layer output matrix:
  
  \begin{equation}\label{eq4}
		\mathbf{H} = \left[
		\begin{array}{c}
			\mathbf{h}(\mathbf{x}_1) \\
			\vdots \\
			\mathbf{h}(\mathbf{x}_N) 
		\end{array}
		\right] 
	\end{equation}
  	where $ \mathbf{h}(\mathbf{x}) = \left[h_1(\mathbf{x}), h_2(\mathbf{x}), \ldots, h_m(\mathbf{x})\right]$ is a~nonlinear feature mapping from $n$-dimensional input space to $ m $-dimensional feature space, and $ h_j(\mathbf{x}) $ is an activation function 
	of the $ j $-th node (a sigmoid in our case).  
	
	\item Calculate the output weights:
	\begin{equation}\label{eq5}
		\boldsymbol{\beta} = \mathbf{H}^+\mathbf{Y}
	\end{equation}
	where $ \boldsymbol{\beta} \in \mathbb{R}^{m\times n}$ is a~matrix of output weights, $ \mathbf{Y} \in \mathbb{R}^{N\times n}$ is a~matrix of target output patterns, and $ \mathbf{H}^+ \in \mathbb{R}^{m\times N} $ is the Moore-Penrose generalized inverse of matrix $ \mathbf{H} $.

\end{enumerate}

Typically, the hidden node weights and biases are i.i.d random variables both generated from the same symmetrical interval $ a_{j,i}, b_j \sim U(-u, u) $. It was pointed out in \cite{Dud19} and \cite{Dud20a} that as the weights and biases have different functions they should be selected separately. The weights decide about the sigmoid slopes and should reflect the TF complexity, while the biases decide about the sigmoid shift and should ensure the placement of the most nonlinear sigmoid fragments, i.e. the fragments around the sigmoid inflection points, into the input hypercube. These fragments, unlike saturation fragments, are most useful for modeling TF fluctuations.

Recently, to improve the performance of randomized FNNs, several new methods of generating the hidden node parameters have been proposed. Among them is the random $a$ method (R$a$M) which was proposed in \cite{Dud20a}. In the first step, this method randomly selects weights from the interval whose bounds $u$ are adjusted to the TF complexity, $ a_{j,i} \sim U(-u, u) $. Then, to ensure the introduction of the sigmoid inflection points into the input hypercube, the biases are calculated from:

\begin{equation}\label{eq6}
b_j = -\mathbf{a}_j^T\mathbf{x}^*_j
\end{equation}
where $\mathbf{x}^*_j$ is one of the training x-patterns selected for the $j$-th hidden node at random.

The second method proposed in \cite{Dud20a}, called the random $\alpha$ method (R$\alpha$M), instead of generating weights, generates the slope angles of sigmoids. This changes the distribution of weights, which typically is a uniform one. This new distribution ensures that the slope angles of sigmoids are uniformly distributed, and so improves results by preventing overfitting, especially for highly nonlinear TFs. This method, in the first step, selects randomly the slope angles of the sigmoids, $ \alpha_{j,i} \sim U(\alpha_{\min}, \alpha_{\max}) $. Then, the the weights are calculated from:

\begin{equation}
a_{j,i}=4 \tan \alpha_{j,i} 
\label{eq7}
\end{equation}

Finally, the biases are determined from \eqref{eq6}. To simplify the optimization process, the lower bound for the angles, $\alpha_{\min}$, can be set as $0^\circ$. In such a case only one parameter decides about the model flexibility, i.e. $\alpha_{\max} \in (0^\circ, 90^\circ)$. This is what we used in our simulation study.   

To improve further FNN randomized learning, a data-driven method (DDM) was proposed in \cite{Dud20}. This method introduces the sigmoids into randomly selected regions of the input space and adjusts the sigmoid slopes to the TF slopes in these regions. As a result, the sigmoids mimic the TF locally, and their linear combination approximates smoothly the entire TF. In the first step, DDM selects the input space regions by selecting randomly the set of training points, $\{\mathbf{x}^*_j\}_{j=1}^m$. Then, the hyperplanes are fitted to the TF locally in the neighbourhoods of all points  $\mathbf{x}^*_j$. The neighborhood of point $\mathbf{x}^*_j$, $\Psi(\mathbf{x}^*_j)$, contains this point and its $k$ nearest neighbors in $\Phi$. The weights are determined based on the hyperplane coefficients from:

\begin{equation}
a_{j,i} = 4a_{j,i}'
\label{eq8}
\end{equation}
where $a_{j,i}'$ are the coefficients of the hyperplane fitted to neighbourhood $\Psi(\mathbf{x}^*_j)$.

The hidden node biases are calculated from \eqref{eq6}. 

Note that the biases in the above-described approaches are determined based on the weights selected first and the data points. Unlike in the standard approach, they are not chosen randomly from the same interval as the weights. Randomized FNN has two hyperparameters to adjust: number of hidden nodes $m$, and the smoothing parameter, i.e. $u$, $\alpha_{\max}$ or $k$, depending on the method of generating parameters chosen. These hyperparameters decide about the fitting performance of the model and its bias-variance tradeoff. Their optimal values should be selected by cross-validation for a given forecasting problem.  

\section{Simulation Study}

In this section, we apply the proposed randomization-based neural models to forecasting
hourly TS with three seasonalities: yearly, weekly and daily. These TS express electricity demand for four European countries: Poland (PL), Great Britain (GB), France (FR) and Germany (DE).  
We use real-world data collected from \url{www.entsoe.eu}. The data period covers the 4 years from 2012 to 2015. Atypical days such as public holidays were excluded from these data (between 10 and 20 days a year). The forecast horizon $\tau$ is one day, i.e. 24 hours. We forecast the daily load profile for each day of 2015. For each forecasted day, a new training set is created and a new randomized model is optimized and trained. The results presented below are averaged over 100 independent training sessions.

The hyperparameters of randomized FNNs were selected using grid search and 5-fold cross-validation. The number of hidden nodes was selected from the set $\{5, 10, ..., 50\}$. The bounds for weights in R$a$M were selected from $\{0.02, 0.04, ..., \\ 0.2, 0.4, ..., 1\}$. The $\alpha_{\max}$ in R$\alpha$M was selected from $\{2^\circ, 4^\circ, ..., 40^\circ, 45^\circ, ..., 90^\circ\}$. The number of nearest neighbors in DDM was selected from $\{25, 27, ..., 69\}$.  

For comparison, we applied a multilayer perceptron (MLP) for the same forecasting problems. MLP was composed of a single hidden layer with $m$ sigmoid nodes whose number was selected using 5-fold cross-validation from the set $\{2, 4, ..., 24\}$. MLP was trained using Levenberg-Marquardt backpropagation with early stopping to avoid overtraining (20\% of training samples were used as validation samples).

Forecasting quality metrics for the test data are presented in Table~\ref{tab1}. They include: mean absolute percentage error ($\mape$), median of $\ape$, root mean square error ($\rmse$), mean percentage error ($\mpe$), and standard deviation of percentage error ($\pe$) as a measure of the forecast dispersion.

\begin{table}[h]
\caption{Forecasting results.}
\label{tab1}
\begin{center}
\setlength{\tabcolsep}{8pt}
\begin{tabular}{@{}llrrrr@{}}
\toprule
& & R$a$M & R$\alpha$M & DDM & MLP \\ 
\midrule
\textbf{PL data}&$\mape$	&	1.32	&	1.32	&	1.35	&	1.37	\\
&Median($\ape$)	&	0.93	&	0.94	&	0.94	&	0.96	\\ 
&$\rmse$	&	358.86	&	364.13	&	380.77	&	374.86	\\
&$\mpe$	&	0.40	&	0.39	&	0.39	&	0.26	\\
&Std($\pe$)	&	1.94	&	1.98	&	2.09	&	2.07	\\ 
\midrule
\textbf{GB data}&$\mape$	&	2.61	&	2.62	&	2.80	&	2.93	\\
&Median($\ape$)	&	1.88	&	1.90	&	1.99	&	2.17	\\ 
&$\rmse$	&	1187.60	&	1184.58	&	1382.97	&	1314.78	\\
&$\mpe$	&	-0.61	&	-0.61	&	-0.58	&	-0.60	\\
&Std($\pe$)	&	3.57	&	3.56	&	4.16	&	3.99	\\ 
\midrule
\textbf{FR data}&$\mape$	&	1.67	&	1.69	&	1.81	&	1.87	\\
&Median($\ape$)	&	1.15	&	1.16	&	1.25	&	1.31	\\ 
&$\rmse$	&	1422.60	&	1433.90	&	1530.15	&	1565.70	\\
&$\mpe$	&	-0.42	&	-0.39	&	-0.45	&	-0.39	\\
&Std($\pe$)	&	2.60	&	2.61	&	2.78	&	2.85	\\ 
\midrule
\textbf{DE data}&$\mape$	&	1.38	&	1.39	&	1.43	&	1.58	\\
&Median($\ape$)	&	0.96	&	0.98	&	0.99	&	1.09	\\ 
&$\rmse$	&	1281.14	&	1242.36	&	1333.79	&	1452.54	\\
&$\mpe$	&	0.14	&	0.14	&	0.10	&	0.04	\\
&Std($\pe$)	&	2.22	&	2.13	&	2.34	&	2.50	\\ 
\bottomrule
\end{tabular}
\end{center}
\end{table}

\begin{figure}[h]
\centering
\includegraphics[width=0.23\textwidth]{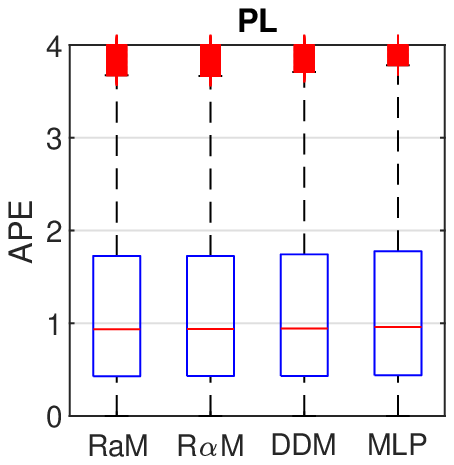}
\includegraphics[width=0.23\textwidth]{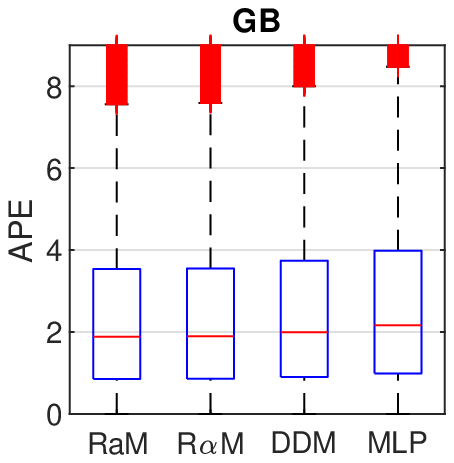}
\includegraphics[width=0.23\textwidth]{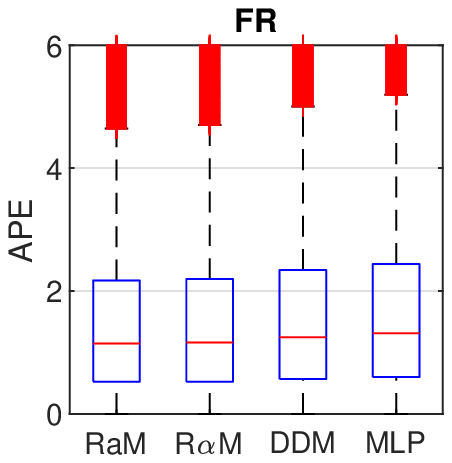}
\includegraphics[width=0.23\textwidth]{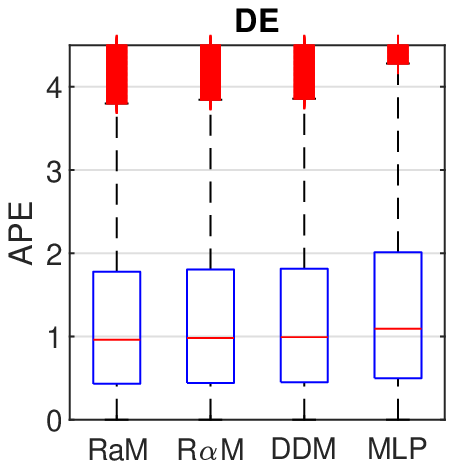}
\caption{Boxplots of $\ape$.} \label{figbx}
\end{figure}

\begin{figure}[h]
\centering
\includegraphics[width=0.65\textwidth]{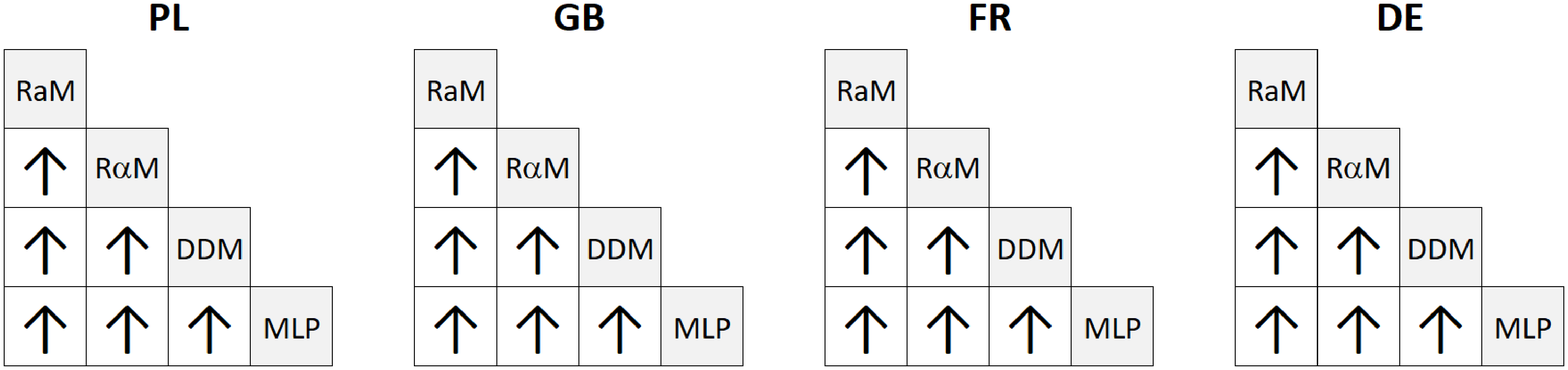}
\caption{Results of the Wilcoxon signed-rank test for $\ape$.} \label{fig4}
\end{figure}

More detailed results, i.e. distributions of $\ape$, are shown in Fig. \ref{figbx}. Based on $\ape$, we performed a Wilcoxon signed-rank test with $\alpha=0.05$ to indicate the most accurate models.     
Fig. \ref{fig4} depicts pairwise comparisons of the models. The arrow lying at the intersection of the two models indicates which of them gave the significantly lower error. A lack of an arrow means that both models gave statistically indistinguishable errors. 

As can be seen from Table \ref{tab1} and Fig. \ref{fig4}, the randomization-based FNNs gave significantly lower errors than fully-trained MLP for each dataset. According to the Wilcoxon test, R$a$M outperformed the other approaches.  

$\mpe$ shown in Table \ref{tab1} allows us to asses the bias of the forecasts produced by different models. A positive value of $\mpe$ indicates underprediction, while a negative value indicates overprediction. As can be seen from Table \ref{tab1}, for PL and DE data the bias was positive, whilst for GB and FR data it was negative. The forecasts produced by MLP for PL and DE were less biased than the forecast produced by randomized FNNs. 

\begin{figure}[h]
\centering
\includegraphics[width=0.243\textwidth]{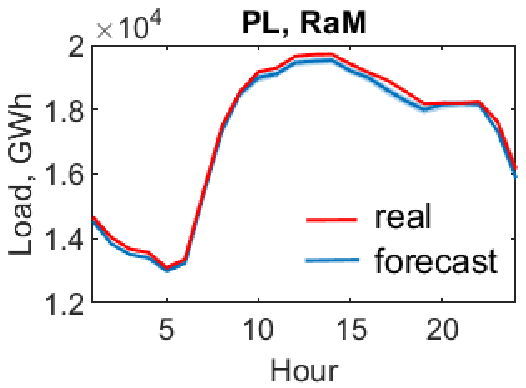}
\includegraphics[width=0.243\textwidth]{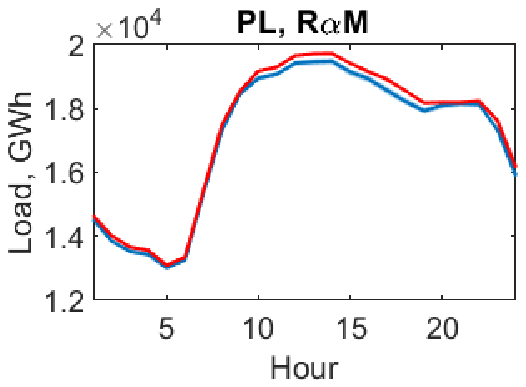}
\includegraphics[width=0.243\textwidth]{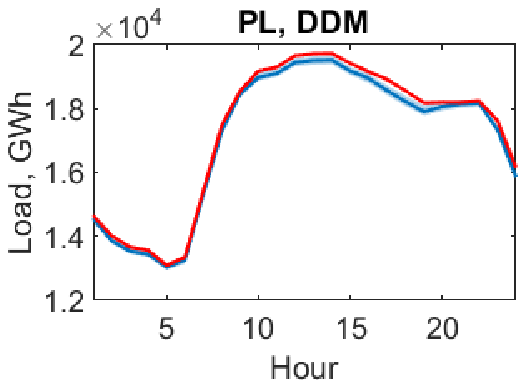}
\includegraphics[width=0.243\textwidth]{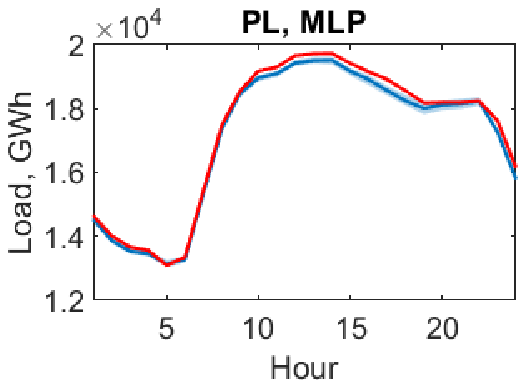}
\includegraphics[width=0.243\textwidth]{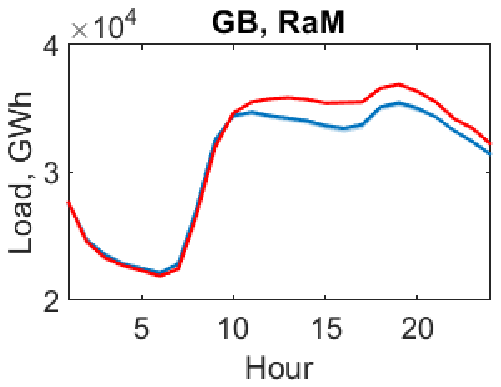}
\includegraphics[width=0.243\textwidth]{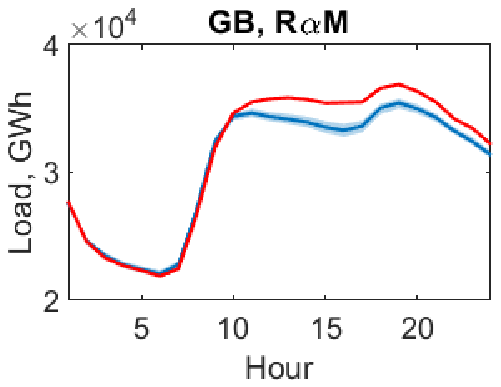}
\includegraphics[width=0.243\textwidth]{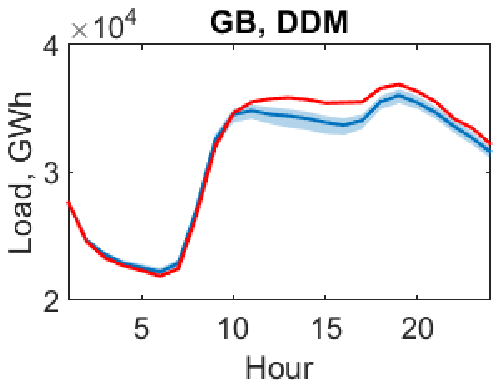}
\includegraphics[width=0.243\textwidth]{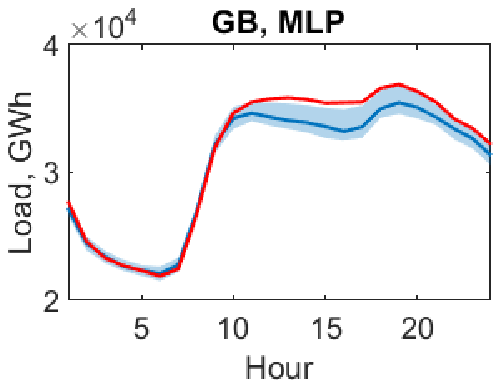}
\includegraphics[width=0.243\textwidth]{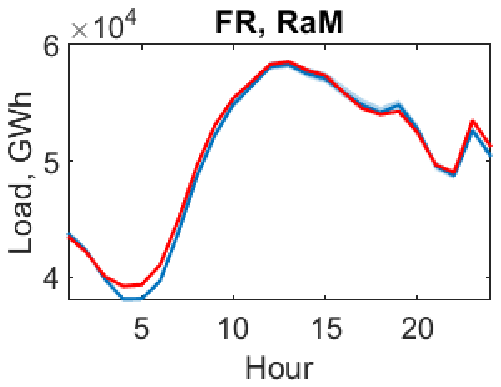}
\includegraphics[width=0.243\textwidth]{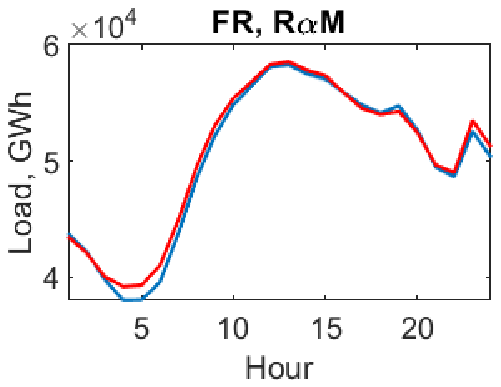}
\includegraphics[width=0.243\textwidth]{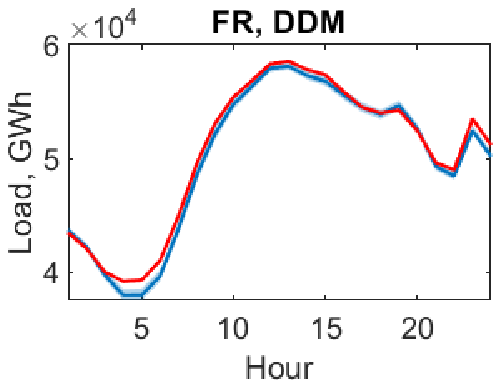}
\includegraphics[width=0.243\textwidth]{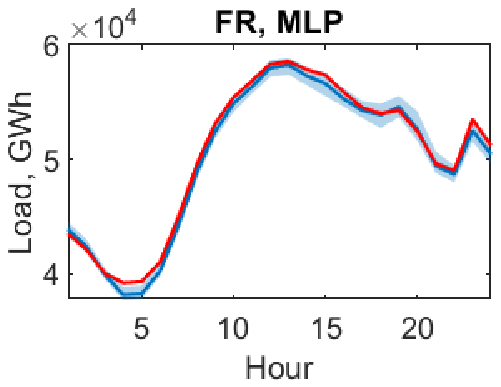}
\includegraphics[width=0.243\textwidth]{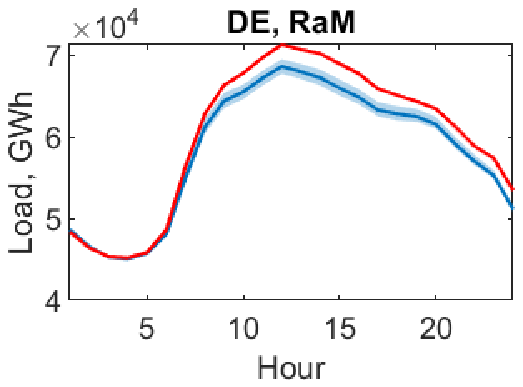}
\includegraphics[width=0.243\textwidth]{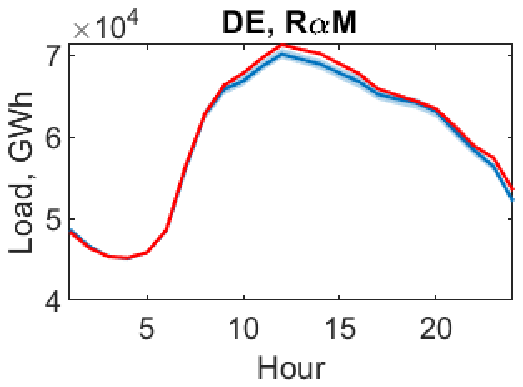}
\includegraphics[width=0.243\textwidth]{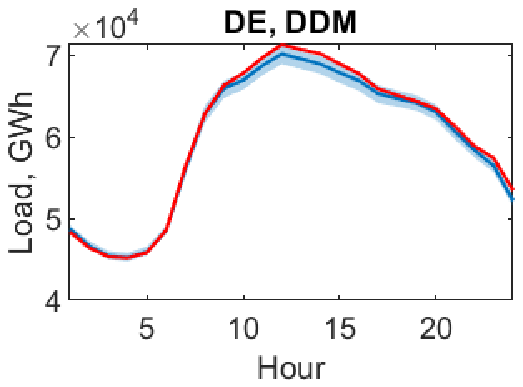}
\includegraphics[width=0.243\textwidth]{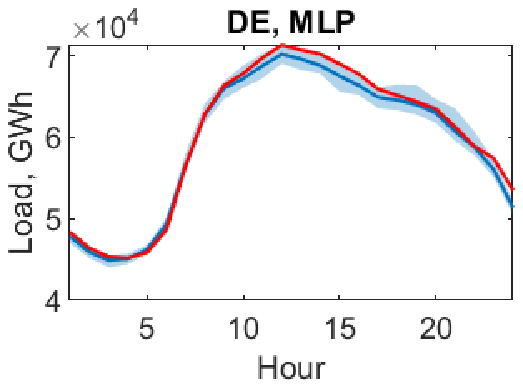}
\caption{Examples of forecasts (shaded regions are  5th  and  95th  percentiles,  measured  over  100  trials).}
\label{figff}
\end{figure}

Fig. \ref{figff} presents examples of forecasts of the daily load profiles produced by the examined models. Note that the proposed models generate multi-output response, maintaining the relationships between the output variables (y-pattern components). In the case of single-output models, these relationships are ignored because the variables are predicted independently. This may cause a lack of smoothness in the forecasted curve (zigzag effect; see for example \cite{Dud18}).

\begin{figure}[h]
\centering
\includegraphics[width=0.243\textwidth]{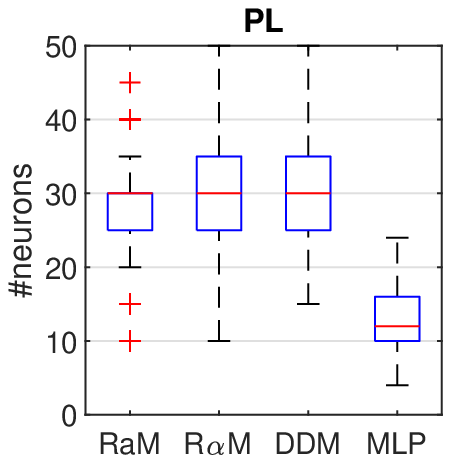}
\includegraphics[width=0.243\textwidth]{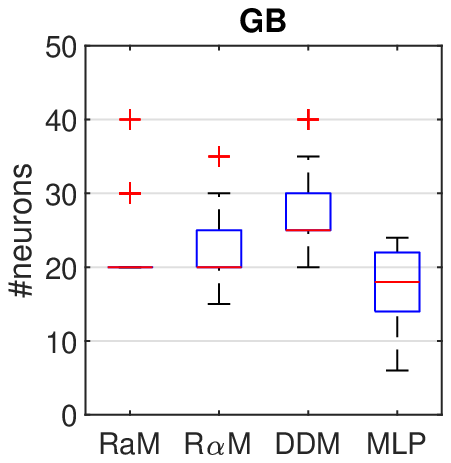}
\includegraphics[width=0.243\textwidth]{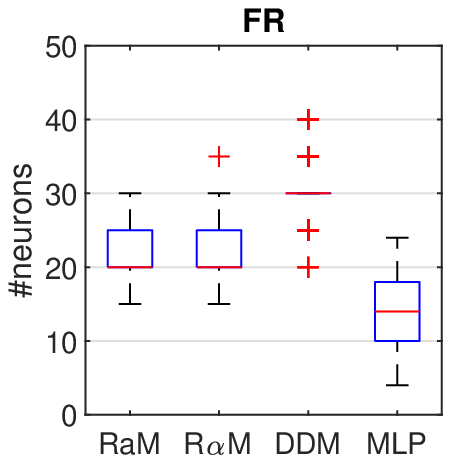}
\includegraphics[width=0.243\textwidth]{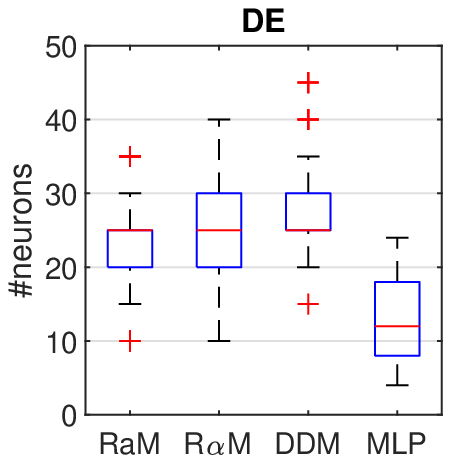}
\caption{Boxplots of the optimal number of hidden nodes.} \label{fighn}
\end{figure}

Fig. \ref{fighn} shows the optimal numbers of hidden nodes selected in the cross-validation procedure. Obviously, the number of hidden nodes is dependent on TF complexity.   
The forecasting problem for PL required  the greatest number of nodes for randomized FNNs, around 30, regardless of the learning method. MLP for PL needed many fewer nodes, 12 on average. Other forecasting problems were solved by randomized FNNs with fewer hidden nodes, from 20 to 30 on average. For these problems, the difference in the number of nodes between MLP and randomized FNNs was not as large as for PL data. The relatively small number of hidden nodes in randomized FNNs (note that randomized learning usually requires hundreds or even thousands of nodes) results from TS representation by unified patterns and the decomposition of the forecasting problem (a separate model for each forecasting task, i.e. every day in 2015, trained on the selected patterns).

\begin{figure}[h]
\centering
\includegraphics[width=0.243\textwidth]{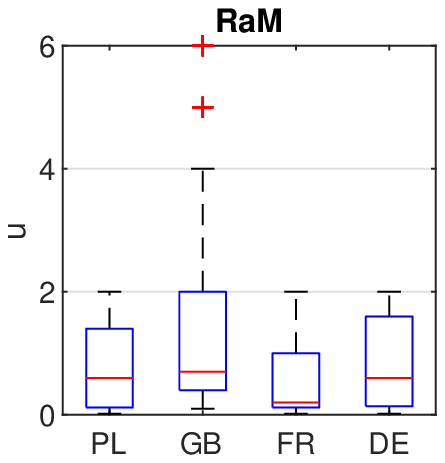}
\includegraphics[width=0.243\textwidth]{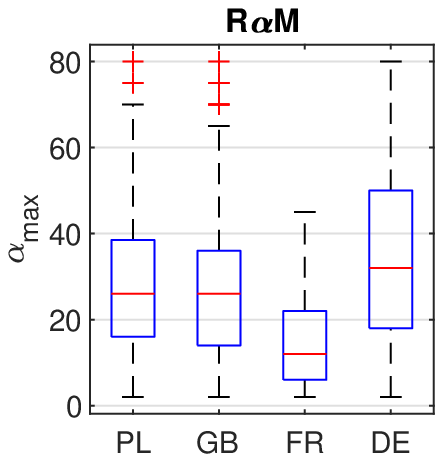}
\includegraphics[width=0.243\textwidth]{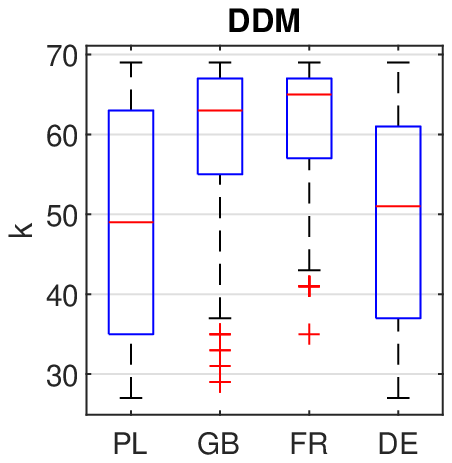}
\caption{Boxplots of the optimal smoothing parameters.} \label{figsp}
\end{figure}

The optimal values of smoothing parameters for the randomized learning methods are depicted in Fig. \ref{figsp}. As can be seen from this figure, the optimal value of the bound for weights in R$a$M varies from 0.2 for FR to 0.7 for GB on average, which correspond to sigmoid slope angles from around $3^\circ$ to $10^\circ$ (see \cite{Dud20a}). The optimal value of the bound for slope angle in R$\alpha$M varies from $12^\circ$ for FR to $32^\circ$ for DE on average. Note also the high value of $k$ in DDM (from $49$ for PL to $65$ for FR on average) in relation to the number of training points, which ranged from 150 to 200. Thus, for our forecasting problems we can expect flat TFs without fluctuations. Such TFs can be modeled using R$a$M. Its competitors, R$\alpha$M and DDM, reveal their strengths in modeling highly nonlinear TFs with fluctuations (see \cite{Dud20a}, \cite{Dud20}).  

\section{Conclusion}

Forecasting TS with multiple seasonality is a challenging problem, which we propose to solve with randomized FNNs. 
Unlike fully-trained FNNs, randomized FNNs learn extremely fast and are easy to implement. The simulation study showed that their forecasting accuracy is comparable to the accuracy of fully-trained NNs. To deal with nonstationary TS with multiple seasonal periods, the proposed approach employs a pattern representation of the TS. This representation simplifies the relationship between input and output data and makes the problem easier to solve using simple regression models. 

The effectiveness of the randomized FNNs in modeling nonlinear target functions was achieved due to the application of new methods of generating hidden node parameters. These methods, using different approaches, introduce the steepest fragments of sigmoids, which are most useful for modeling TF fluctuations, into the input hypercube and adjust their slopes to TF complexity. This makes the model more flexible, more data-dependent, and more dependent on the complexity of the solved forecasting problem.

In a future study, we plan to introduce an attention mechanism into our randomization-based forecasting models to select training data and develop an ensemble approach for these models. 

%
%
%

\end{document}